\documentclass{article}

\usepackage{arxiv}
\usepackage{amsthm}
\usepackage[utf8]{inputenc} 
\usepackage[T1]{fontenc}    
\usepackage{hyperref}       
\usepackage{url}            
\usepackage{booktabs}       
\usepackage{amsfonts}       
\usepackage{nicefrac}       
\usepackage{microtype}      
\usepackage{lipsum}
\usepackage{amssymb,amsmath,amsfonts}
\usepackage{latexsym}
\usepackage{graphicx}
\usepackage{mathrsfs}
\usepackage{geometry}
\usepackage{fancyhdr}
\usepackage{booktabs}
\usepackage{amsmath}
\usepackage{amsfonts}
\usepackage{amssymb}
\usepackage{multirow}
\usepackage{array}
\usepackage{bm}

\def\x{\mathbf x}

\def\s{\mathbf s}

\def\W{\mathbf W}

\def\L{\mathcal{L}}

\DeclareMathOperator{\sign}{sign} 
\DeclareMathOperator{\BN}{BN}
\DeclareMathOperator{\var}{\mathbb{V}}

\def \real{\rm I\!R}

\def \L {\mathcal{L}}

\DeclareMathOperator{\tern}{tern}

\begin{document}




\title{Batch Normalization in Quantized Networks}
		
\author{Eyy\"ub Sari \\ eyyub.sari@huawei.com  \\ Huawei Noah's Ark Lab \And Vahid Partovi Nia\thanks{Corresponding author.} \\ vahid.partovinia@huawei.com  \\ Huawei Noah's Ark Lab}

		
\maketitle
	\begin{abstract}
	    Implementation of quantized neural networks on computing hardware leads to considerable speed up and memory saving. However, quantized deep networks are difficult to train and batch~normalization (BatchNorm) layer plays an important role in training full-precision and quantized networks.  Most studies on BatchNorm are focused on full-precision networks, and there is little research in understanding  BatchNorm affect in quantized training which we address here. We show  BatchNorm  avoids gradient explosion which is counter-intuitive and recently observed in numerical experiments by other researchers.
	\end{abstract}

\section{Introduction}
Deep Neural Networks (DNNs) compression through quantization is a recent direction in edge implementation of deep networks. Quantized networks are simple to deploy on hardware devices with constrained resources such as cell phones and IoT equipment.  Quantized networks not only consume less memory and simplify computation, it also yields energy saving.
Two well-known extreme quantization schemes are binary (one bit) and and ternary (two bit) networks, which  allow up to $32\times$ and $16\times$ computation speed up, respectively. Binary quantization  only keep track of the sign  $\{-1, +1\}$ and ignores the magnitude, and ternary quantization  extends the binary case to $\{-1, 0, +1\}$ to allow for sparse representation. 
BatchNorm facilitates  neural networks training as a known fact. A common intuition suggests  BatchNorm  matches input and output first and second  moments.  There are two other clues among others: \cite{ioffe2015batchnorm} claim that BatchNorm corrects covariate shift, and  \cite{santurkar2018bnoptim} show BatchNorm bounds the gradient and makes the optimization smoother in full-precision networks. None of these arguments work for quantized networks! The role of BatchNorm is to prevent exploding gradient  empirically observed in \cite{ardakani2018learningrecbinter} and  \cite{hou2019normalization}.

\section{Full-Precision Network}

Suppose 
a mini batch of size $B$ for a given neuron $k$. Let 
$\hat{\mu}_{k}, \hat{\sigma}_{k}$ be the mean and the standard deviation of the dot product, between inputs and weights, $s_{bk}, b=1,\dots B$. For a given layer $l$, BatchNorm is defined as $
			\BN(s_{bk}) \equiv z_{bk} = \gamma_k \hat{s}_{bk} + \beta_k,$
where $\hat{s}_{bk} = \frac{s_{bk} - \hat{\mu}_{k}}{\hat{\sigma}_{k}} $ is the standardized dot product and the pair $(\gamma_k$,  $\beta_k)$ is trainable, initialized with $(1, 0)$.

Given the objective function $\L(.)$, BatchNorm parameters are trained in backpropagation 
$$	    \frac{\partial \L}{\partial \beta_k} = \sum_{b=1}^{B}\frac{\partial \L}{\partial z_{bk}},
	    \quad
		\frac{\partial \L}{\partial \gamma_k} = \sum_{b=1}^{B} \frac{\partial \L}{\partial z_{bk}} \hat{s}_{bk},
$$
For a given layer $l$, it is easy to prove $\frac{\partial \L}{\partial s_{bk}}$ equals 
    \begin{equation}\label{eq:bnderiv}
		\frac{\gamma_k}{\hat{\sigma}_{k}} \Big(- \frac{1}{B}\sum_{b'=1}^{B} \frac{\partial \L}{\partial z_{b'k}} - \frac{\hat{s}_{bk}}{B}\sum_{b'=1}^{B} \frac{\partial \L}{\partial z_{b'k}} \hat{s}_{b'k} + \frac{\partial \L}{\partial z_{bk}} \Big).
    \end{equation}

Assume weights and activations are  independent, and identically distributed (iid) and centred about zero. Formally, denote the dot product vector $\s_b^l \in {\real}^{K_l}$  of sample $b$ in layer $l$, with $K_l$ neurons. Let $f$ be the element-wise activation  function,  $\x_b$ be the input vector, $\W^l \in {\real}^{K_{l-1} \times K_l}$ with elements $\mathbf W^l = [w^l_{kk'}]$ be the weights matrix; one may  use $w^l$ to denote an identically distributed elements of layer $l$. It is easy to verify 
\begin{eqnarray*}
			\frac{\partial \L}{\partial s^l_{bk}} &=& f'(s^l_{bk})\sum_{k'=1}^{K_{l+1}}w^{l+1}_{kk'}\frac{\partial \L}{\partial s^{l+1}_{bk'}}, \\
			\frac{\partial \L}{\partial w^l_{k'k}}&=&  \sum_{b=1}^Bs^{l-1}_{bk'}\frac{\partial \L}{\partial s^l_{bk}}.
\end{eqnarray*}

Assume that the feature element $x$ and the weight element $w $ are centred and iid. Reserve $k$ to index the current neuron and use $k'$ for the previous or the next layer neuron  and where $\var(w^{l'})$ is the variance of the weight in layer $l'$ 
				$\var(s^l_{bk}) = \var(x)\prod_{l'=1}^{l-1} K_{l'} \var(w^{l'}),$
$$			\var(\frac{\partial \L}{\partial s^l_{bk}}) = \var(\frac{\partial \L}{\partial s^L})\prod_{l'=l+1}^LK_{l'}\var(w^{l'}),
			$$
which explodes or vanishes depending on $\var(w^{l'})$. This is the main reason common full-precision initialization methods suggest $\var(w^l)= {1\over K_{l}}$.
For any full-precision network,  BatchNorm affects backpropagation as 
		\begin{eqnarray}
		\var \Big(\frac{\partial \L}{\partial s^l_{bk}} \Big) 
			&=& \Big(\frac{\gamma^{l}_k}{B {\hat{\sigma}^l}_{k}} \Big)^2 \{B^2 + 2B - 1 + \var(\hat{s}_{bk}^{l^2}) \} \nonumber \\ 
			&&K_{l+1}\var(w^{l+1})\var \Big(\frac{\partial \L}{\partial s^{l+1}} \Big).
			\label{eq:backbnnvar}
		\end{eqnarray}		
\section{Binary Network}
Controlling the variance has no fundamental effect on forward propagation if $s_{bk}$ is symmetric about zero as the sign function filters the magnitude and only keeps the sign of the dot product. The term $b_k= \mu_{k} - \frac{\hat{\sigma}_{k}}{\gamma_k}\beta_k$ can be regarded as as a new trainable parameter, thus BatchNorm layer can be replaced by adding biases to the network to compensate.  \cite{sari2019study} shows that  the gradient variance for binary quantized networks  without BatchNorm is
$$
		\var \Big(\frac{\partial \L}{\partial s^l_{bk}} \Big)  = \var\Big(\frac{\partial \L}{\partial s^L}\Big)\prod_{l'=l+1}^L K_{l'},$$
and with BatchNorm is  
$$ \var \Big(\frac{\partial \L}{\partial s^l_{bk}} \Big)  =\prod_{l'=l}^{L-1}\frac{K_{l'+1}}{K_{l'-1}}\var\Big(\frac{\partial \L}{\partial s^{L}}\Big)+o\left({1\over B^{1-\epsilon}} \right),
$$ for an arbitrary $0<\epsilon<1$.

Gradients are stabilized only if $\Big(\frac{\gamma^{l}_k}{B} \Big)^2 \{B^2 + 2B - 1 + \var(\hat{s}_{bk}^{l^2}) \} \approx 1$. Moving from full-precision weight $w$ to binary weight $\tilde w = \sign (w)$ changes the situation dramatically:  i)  BatchNorm corrects exploding gradients in BNNs as the layer width ratio  ${K_{l+1}\over K_{l-1}}\approx 1$ in common neural models.  If this ratio diverges from unity binary training is problematic even with  BatchNorm.

\section{Ternary Network}
Ternary neural networks (TNNs) are studied in \cite{Sari_Nia_2020} and the BatchNorm effect is detailed there. Full-precision weights during training  are ternarized during forward propagation. Given a threshold $\Delta$  ternary quantization function is
\begin{equation}
    \tern (x) = \begin{cases}
-1~&\text{ if } x < -\Delta \\
+1~&\text{ if } x > \Delta \\
0~&\text{ if } -\Delta \leq x \leq \Delta \\
\end{cases}
\end{equation}
Let's suppose the threshold is given so that the learning is feasible, for instance $\Delta$ is tuned so that $<50\%$ of ternary weights are set to zero \begin{equation} \label{eq:varter}
\var(\tilde w^l_t) = 2p_1 = 1 - \frac{\Delta}{\sqrt{\frac{6}{K_l}}}.    
\end{equation}
In the literature  \cite{li2016twn} suggests to set $\Delta_l = 0.7\mathbb{E}(|w^l|)$. Under simplified assumptions of iid weight and activation 
\begin{equation}\label{eq:delta}
    \Delta_l = \frac{0.7}{2}\sqrt{\frac{6}{K_l}}
\end{equation}
and \eqref{eq:varter} reduces to $\var(\tilde w^l_t) = 1-\frac{0.7}{2} = 0.65$. In this setting, variance is bigger than $2\over K_l$ which produces exploding gradients similar to the binary case.
Suppose weights and activation are iid and weights are centred about zero, for a layer $l$,
\begin{equation}
     {\hat{\sigma}^2}_{k} = K_{l-1}\frac{1}{2}\var(\hat{s}^{l-1}_b)\var(\tilde w_t^{l}) = K_{l-1}\frac{1}{2}\var(\tilde w_t^{l}).
\end{equation}
Therefore \eqref{eq:backbnnvar} reduces to
\begin{eqnarray}
\label{eq:varbackbnter}
    \var \Big(\frac{\partial \L}{\partial s^l_{bk}} \Big) 
			&=& \left\{1 + o\left({1\over B^{1-\epsilon}}\right)\right\}\\ && \frac{K_{l+1}}{K_{l-1}}\var\Big(\frac{\partial \L}{\partial s^{l+1}} \Big),
\end{eqnarray}
see \cite{Sari_Nia_2020} for details. Similar to the binary case, in most deep architectures $K_{l+1} \approx K_{l-1}$ or equivalently  $\frac{K_{l+1}}{K_{l-1}} \approx 1$, so the variance would not explode for networks with BatchNorm layer.
\section{Conclusion}
We derived the analytical expression for full-precision network under assumptions of \cite{he2015init} and extended it for binary and ternary case. Our study shows that the real effect of BatchNorm is played in scaling. The main role of BatchNorm in quantized training is to adjust gradient explosion.

\bibliographystyle{plain}
\bibliography{batchnorm.bib}
\end{document}